%% file: main.tex
\documentclass[10pt,journal,compsoc]{IEEEtran}

\ifCLASSOPTIONcompsoc
  \usepackage[nocompress]{cite}
\else
  \usepackage{cite}
\fi
\ifCLASSINFOpdf
\else
\fi
\usepackage{graphicx}
\usepackage{xcolor}
\usepackage{color}
\usepackage{mathrsfs}
\usepackage{amsmath}
\usepackage{amssymb}
\usepackage{ragged2e}
\usepackage{marvosym}
\usepackage{algorithm}
\usepackage{listings}
\usepackage{algorithmic}
\usepackage{booktabs}

\def\eg{\emph{e.g.}}

\def\etc{{\em etc.}} 
\definecolor{dkgreen}{rgb}{0,0.6,0}
\definecolor{gray}{rgb}{0.5,0.5,0.5}
\definecolor{mauve}{rgb}{0.58,0,0.82}

\newcommand{\ours}{Fast-BEV }

\usepackage{url}
\usepackage{booktabs}
\usepackage{array, caption, threeparttable}
\usepackage[font=small,labelfont=bf]{caption}
\begin{document}
\title{Fast-BEV: A Fast and Strong Bird’s-Eye View Perception Baseline}
\author{
Yangguang Li, Bin Huang, Zeren Chen, Yufeng Cui, Feng Liang, Mingzhu Shen, \\Fenggang Liu, Enze Xie, Lu Sheng\textsuperscript{\Letter}, Wanli Ouyang, Jing Shao
\IEEEcompsocitemizethanks{

\IEEEcompsocthanksitem Y. Li and F. Liu are with SenseTime Research.
\IEEEcompsocthanksitem B. Huang is with Hozon New Energy Automobile Co., Ltd.
\IEEEcompsocthanksitem Z. Chen, Y. Cui, M. Shen and L. Sheng are with Beihang University.
\IEEEcompsocthanksitem W. Ouyang and J. Shao are with Shanghai AI Laboratory.
\IEEEcompsocthanksitem F. Liang is with The University of Texas at Austin.
\IEEEcompsocthanksitem E. Xie is with The University of Hong Kong.
\IEEEcompsocthanksitem \textsuperscript{\Letter}: Corresponding author. 
}
}

\markboth{Fast-BEV: A Fast and Strong Bird’s-Eye View Perception Baseline}
{Shell \MakeLowercase{\textit{et al.}}: Bare Demo of IEEEtran.cls for Computer Society Journals}
\IEEEtitleabstractindextext{
\begin{abstract}
    \justifying
    Recently, perception task based on Bird’s-Eye View (BEV) representation has drawn more and more attention, and BEV representation is promising as the foundation for next-generation Autonomous Vehicle (AV) perception.
    However, most existing BEV solutions either require considerable resources to execute on-vehicle inference or suffer from modest performance.
    This paper proposes a simple yet effective framework, termed \ours, which is capable of performing faster BEV perception on the on-vehicle chips.
    Towards this goal, we first empirically find that the BEV representation can be sufficiently powerful without expensive transformer based transformation or depth representation.
    Our \ours consists of five parts, we innovatively propose (1) a lightweight deployment-friendly view transformation which fast transfers 2D image features to 3D voxel space, (2) a multi-scale image encoder which leverages multi-scale information for better performance, (3) an efficient BEV encoder which is particularly designed to speed up on-vehicle inference. We further introduce (4) a strong data augmentation strategy for both image and BEV space to avoid over-fitting, (5) a multi-frame feature fusion mechanism to leverage the temporal information.
    Among them, (1) and (3) enable \ours to be fast inference and deployment friendly on the on-vehicle chips, (2), (4) and (5) ensure that \ours has competitive performance. All these make \ours a solution with high performance, fast inference speed, and deployment-friendly on the on-vehicle chips of autonomous driving.  
    Through experiments, on 2080Ti platform, our R50 model can run 52.6 FPS with 47.3\% NDS on the nuScenes validation set, exceeding the 41.3 FPS and 47.5\% NDS of the BEVDepth-R50 model\cite{li2022bevdepth} and 30.2 FPS and 45.7\% NDS of the BEVDet4D-R50 model\cite{huang2022bevdet4d}.
    Our largest model (R101@900x1600) establishes a competitive 53.5\% NDS on the nuScenes validation set. 
    We further develop a benchmark with considerable accuracy and efficiency on current popular on-vehicle chips.
    The code is released at: \url{https://github.com/Sense-GVT/Fast-BEV}.

\end{abstract}

}

\maketitle

\IEEEdisplaynontitleabstractindextext

\IEEEpeerreviewmaketitle

\input{Sections/introduction.tex}

\input{Sections/related_work.tex}
\input{Sections/method.tex}

\input{Sections/experiments.tex}
\input{Sections/benchmark.tex}
\input{Sections/application_discussion.tex}

\input{Sections/conclusion.tex}

\section*{Acknowledgments}
This work is supported in part by National Key Research and Development Program of China (No. 2021YFB1714300), and National Natural Science Foundation of China (62132001).

\ifCLASSOPTIONcaptionsoff
  \newpage
\fi
\bibliographystyle{IEEEtran}
\bibliography{egbib.bib}

\end{document}

%% file: Sections/introduction.tex
\section{Introduction}
A fast and accurate 3D perception system is essential for autonomous driving.
Classic methods~\cite{zhou2018voxelnet,lang2019pointpillars,shi2019pointrcnn} rely on the accurate 3D information provided by LiDAR point clouds. 
However, LiDAR sensors usually cost thousands of dollars~\cite{neuvition_2022}, hindering their applications on economical vehicles. 
Pure camera-based Bird’s-Eye View (BEV) methods~\cite{wang2022detr3d, xie2022m, huang2021bevdet,liu2022petr, li2022bevformer,li2022bevdepth,liu2022bevfusion} have recently shown great potential for their impressive 3D perception capability and economical cost.
They basically follow the paradigm: multi-camera 2D image features are transformed to the 3D BEV feature in ego-car coordinates, then specific heads are applied on the unified BEV representation to perform specific 3D tasks, \eg, 3D detection, segmentation, \etc. 
The unified BEV representation can handle a single task separately or multi-tasks simultaneously, which is highly efficient and flexible.

\begin{figure*}[t]
    \begin{center}
    \includegraphics[width=1\linewidth]{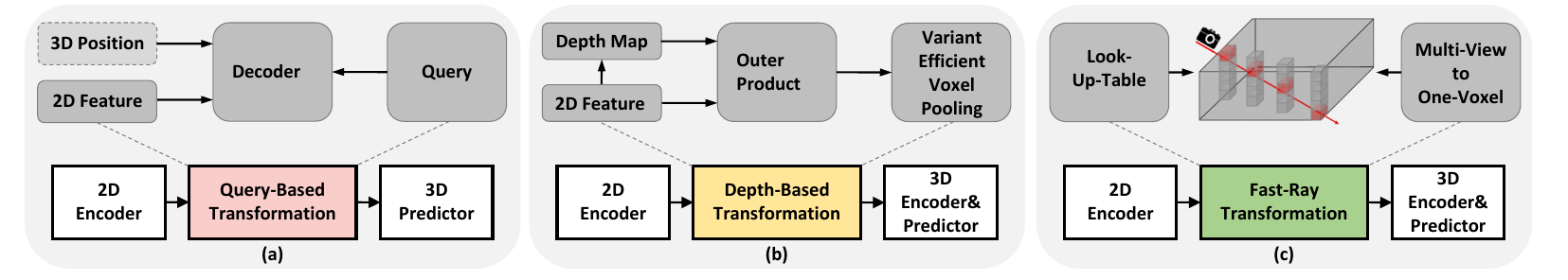}
    \end{center}
    \caption{Methods comparison of view transformation. (a) Query-Based Transformation: methods with transformer's attention mechanism. (b) Depth-Based Transformation: methods with depth distribution prediction. (c) \textit Fast-Ray Transformation (Ours): Uniform depth distribution along the camera ray with Look-Up-Table and Multi-View to One-Voxel operations.
    }
\label{fig:method_comparison}
\end{figure*}

To perform 3D perception from 2D image features, state-of-the-art BEV methods on nuScenes~\cite{caesar2020nuscenes} either uses query based transformation~\cite{wang2022detr3d,li2022bevformer} or implicit/explicit depth based transformation~\cite{philion2020lift,huang2021bevdet,li2022bevdepth}.
However, they are difficult to deploy on the on-vehicle chips and suffer from slow inference: (1) Methods based on query-based transformation are shown in Fig.\ref{fig:method_comparison}(a). 
Because the decoder requires the attention mechanism in the transformer for calculation, the actual deployed chip needs to be able to support the attention operator (involving matrix multiplication and MLP layers), so these methods usually require dedicated chips to support.
(2) Methods based on depth-based transformation are shown in Fig.\ref{fig:method_comparison}(b). These methods often require accelerated speed unfriendly voxel pooling operation, and even multi-threaded CUDA cores may not be the optimal solution. Also, this is inconvenient to run on chips that are resource constrained or not supported by the CUDA accelerated inference library.
Moreover, they are time-consuming in inference, which prevents them from actual deployment.
In this paper, we aim to design a BEV perception framework with friendly deployment, high inference speed, and competitive performance for \emph{on-vehicle chips}. \eg, Xavier, Orin, Tesla T4, etc..

Based on these observations, following the principle of M$^2$BEV~\cite{xie2022m}, which assumes a uniform depth distribution along the camera ray during image-to-BEV(2D-to-3D) view transformation, we propose Fast-Ray transformation, as shown in Fig.\ref{fig:method_comparison}(c), which speeds up BEV transformation to a new level with the help of Look-Up-Table and Multi-View to One-Voxel operation. Based on Fast-Ray transformation, we further propose \ours, a faster and stronger fully convolutional BEV perception framework without expensive view transformer~\cite{wang2022detr3d,li2022bevformer} or depth representation~\cite{huang2021bevdet,huang2022bevdet4d,li2022bevdepth}.
The proposed \ours consists of five parts, Fast-Ray transformation, multi-scale image encoder, efficient BEV encoder, data augmentation, and temporal fusion. These together form a framework that endows \ours with fast inference speed and competitive performance. 

Specifically, (1) we innovatively propose Fast-Ray transformation, a lightweight and deployment-friendly view transformation for fast inference, which obtains BEV representations by projecting multi-view 2D image features to 3D voxels along the camera rays. Furthermore, we propose two operations, Look-Up-Table and Multi-View to One-Voxel, to optimize the process for on-vehicle platforms. (2) The view transformation of existing works is time-consuming, and multi-scale projection operations will have a large time overhead, so it is difficult to use in practice. Based on Fast-Ray transformation, our 2D-to-3D projection has extremely fast speed, making multi-scale image encoder with multi-scale projection operation possible. Specifically, unlike most existing works that use the single-scale features as image encoder output, we use a 3-layer multi-scale feature pyramid network (FPN)~\cite{lin2017feature} structure in the image encoder output part. It is followed by the corresponding 3-level multi-scale projection operation. (3) For the BEV encoder, we use very few naive residual network as the basic BEV encoder. On this basis, three dimension reduction operations are used to accelerate the encoder, which are the "space-to-channel" (S2C)~\cite{xie2022m} operator, the multi-scale concatenation fusion (MSCF) operator, and the multi-frame concatenation fusion (MFCF) operator, respectively. (4) We further introduce a strong data augmentation strategy~\cite{huang2021bevdet,li2022bevdepth}, such as flipping, rotation, resize, \etc, for both image and BEV space. The data augmentation, which is performed in the image space and the BEV space separately, not only avoids over-fitting but also achieves better performance. (5) We further introduce temporal fusion ~\cite{huang2022bevdet4d,li2022bevformer}, which extends the \ours from spatial-only space to spatial-temporal space via introducing the temporal feature fusion module, enabling current key-frame leverage the information from historical frames. Integrating (1) and (3), \ours can achieve faster inference speed on the on-vehicle chips. Further aggregating (2), (4), and (5), \ours can achieve a competitive performance.

BEV perception has frequently updated performance benchmarks in academia, such as NuScenes benchmark~\cite{caesar2020nuscenes}, but it has rarely been studied in terms of industrial applications. For the first time, we develop a benchmark with considerable accuracy and efficiency on current popular on-vehicle chips, from latency to performance between on-vehicle chips with different computing power, which provides a reference for the actual deployment of the BEV solutions.

With its high efficiency and competitive performance, \ours breaks the belief that the existing BEV solutions are difficult to deploy on low-computing chips. Simplicity and efficiency are its key advantages. We hope that this work can provide a simple and powerful baseline for the deployment of future task-agnostic BEV perception schemes on low-computing platforms.
Specificlly, the proposed \ours model family shows great performance and can be easily deployed on on-vehicle platforms.
On the nuScenes~\cite{caesar2020nuscenes} dataset, on 2080Ti platform, our R50 model can run 52.6 FPS with 47.3\% NDS on the nuScenes validation set, exceeding the 41.3 FPS and 47.5\% NDS of the BEVDepth-R50 model\cite{li2022bevdepth} and 30.2 FPS and 45.7\% NDS of the BEVDet4D-R50 model\cite{huang2022bevdet4d}.
Our largest model (R101@900x1600) establishes a competitive 53.5\% NDS on the nuScenes validation set. 

In conclusion, our contributions are summarized as follows:

\begin{itemize}
\item We propose a lightweight and deployment-friendly Fast-Ray transformation,  which has two acceleration designs: pre-computing the projection index and projecting to the same voxel feature. Around this, we also design an efficient BEV encoder to accelerate feature extraction. The holistic design enables \ours to be easily deployed on the on-vehicle chips with fast inference speed. 

\item  We first propose a multi-scale image encoder with multi-scale projection and introduce strong data augmentation and multi-frame temporal fusion on our framework, the effectiveness of these techniques enables \ours to achieve the competitive performance on the challenging nuScenes dataset.

\item We first propose a benchmark on the on-vehicle chips. And to the best of our knowledge, the proposed \ours is the first deployment-oriented work targeted at the challenging fast on-vehicle BEV perception. We hope our work can shed light on the industrial-level, faster, on-vehicle BEV perception.

\end{itemize}

%% file: Sections/related_work.tex
\section{Related Work}
\noindent \textbf{Camera-based Monocular 3D Object Detection.}
Detectors in 3D object detection aim at predicting objects' localization and categories in 3D space, given the input generated by LiDAR or camera sensors.
LiDAR-based methods, e.g., CenterPoint~\cite{yin2021center}, tend to use 3D CNNs to extract spatial features from LiDAR points, and further regress to 3D attributes like 3D centers of objects.
Compared with LiDAR sensor, using only camera images as input is not only cheaper, but also provides richer semantic information.
One practical approach of monocular 3D object detection is to learn 3D bounding boxes based on 3D image features.
M3D-RPN~\cite{brazil2019m3d} proposes a 3D region proposal network and depth-aware convolutional layers to improve 3D scene understanding. 
Following FCOS~\cite{tian2019fcos}, FCOS3D~\cite{wang2021fcos3d} directly predicts 3D bounding boxes for each object by converting 3D targets into the image domain.
PGD~\cite{wang2022probabilistic} uses the relations across the objects and a probabilistic representation to capture depth uncertainty to facilitate depth estimation for 3D object detection. 
DD3D~\cite{park2021pseudo} benefits from depth pre-training and significantly improve end-to-end 3D detection.

\noindent \textbf{Camera-based Surrounding 3D Object Detection.}
Recent advance in some large-scale benchmarks~\cite{caesar2020nuscenes,sun2020scalability}, especially using more surrounding views, further promotes the field of 3D perception.
In the camera-based 3D object detection, the newly-proposed multi-view transformation techniques ~\cite{reading2021categorical, philion2020lift, roddick2018orthographic, rukhovich2022imvoxelnet} reformulate the task into a stereo matching problem, where the surrounding image features are transformed into stereo representation like BEV (Bird's-Eye View) or 3D voxels.
For example, LSS ~\cite{philion2020lift} projects pixel-wise features over a predicted depth distribution to generate camera frustums, and then convert the frustums into BEV grid. 
OFT~\cite{roddick2018orthographic} proposes to generate the voxel representations by projecting the pre-defined voxels onto image features.
BEVDet\cite{huang2021bevdet} applies LSS to the fully convolutional method, and first verifies the performance of explicit BEV. M$^2$BEV\cite{xie2022m} first follows OFT's view transformation route that does not predict depth information, and explores BEV multi-task perception on this basis.
BEVDepth~\cite{li2022bevdepth} further extend LSS ~\cite{philion2020lift} with robust depth supervision and efficient pooling operations.
Another roadmap for view transformation is grid-shaped BEV queries.
DETR3D~\cite{wang2022detr3d} and Graph-DETR3D~\cite{chen2022graph} decode each BEV query into a 3D reference point to sample the correlated 2D features from images for refinement.
BEVFormer~\cite{li2022bevformer} introduces spatial cross-attention in 2D-to-3D transformation, allowing each query can aggregate its related 2D features across camera views.
PETR~\cite{liu2022petr} propose 3D coordinate generation to perceive the 3D position-aware features, avoiding generating 3D reference points.
The success of these work motivate us to extend the surrounding multi-camera detection pipeline effectively and efficiently.
In this work, we find using depth distribution is not necessary in LSS-like methods, and can be removed to further speed up the whole pipeline.

\noindent \textbf{Camera-based Multi-View Temporal Fusion.}
Recently, several camera-based methods have attempted to introduce multi-frame fusion during detection, which has been shown to be effective for velocity estimation and box localization in LiDAR-based detectors~\cite{yin2021center,lang2019pointpillars,yan2018second}.
And BEV, as an intermediate feature that simultaneously combines visual information from multiple cameras, is suitable for temporal alignment.
~\cite{saha2021translating} proposes a dynamics module that uses past spatial BEV features to learn spatiotemporal BEV representation.
BEVDet4D~\cite{huang2021bevdet} extends the BEVDet ~\cite{huang2021bevdet} by aligning the multi-frame features and exploiting spatial correlations in ego-motion.
PETRv2~\cite{liu2022petrv2} directly achieve temporal alignment in 3D space based on the perspective of 3D position embeddings.
BEVFormer~\cite{li2022bevformer} designs a temporal self-attention to recursively fuse the history BEV information, similar to the hidden state in RNN models.
Our work is also inspired by temporal alignment.
Specifically, we apply this technique to further improve the performance while maintaining high efficiency.

%% file: Sections/method.tex
\section{Methods}
In this section, we introduce the details of our \ours, which demonstrates the design process of each component of \ours and its advantages over other methods.
First of all, we rethink the BEV perception paradigm from the 2D-to-3D projection process, and in Sec.\ref{sec:rethink}, we analyze the characteristics of the two schemes, LSS and Transformer, and propose a new 2D-to-3D scheme, Fast-Ray transformation, which makes our \ours successful. 
Following this, we introduce the overall architecture of \ours in Sec.\ref{sec:overall}, revealing all the components of \ours.
Last, we group these components into several parts, namely Fast-Ray view transformation in Sec.\ref{sec:projection}, multi-scale image encoder in Sec.\ref{sec:2d encoder}, efficient BEV encoder in Sec.\ref{sec:3d encoder}, data augmentation in Sec.\ref{sec:aug}, temporal fusion in Sec.\ref{sec:multiframe}, and detection head in Sec.\ref{sec:detection}.

\subsection{Rethink BEV Perception 2D-to-3D Projection}\label{sec:rethink}
The most important aspect of BEV perception is how to transfer 2D features to 3D space. As shown in Fig.\ref{fig:method_comparison}, the query-based methods obtain the 3D BEV feature through the attention mechanism in the transformer. This process can be expressed as the Formula~\ref{eq1}.
\begin{equation} \label{eq1}
\begin{aligned}
    F_{bev}(x,y,z) = Attn(q,k,v) \\
\end{aligned}
\end{equation}
where $q$, $k$, $v$ mean for query, key, value,  $q\subset P_{xyz}$,  $k,v\subset F_{2D}(u,v)$.  $P_{xyz}$ is predefined anchor points in 3D space, and $F_{2D}(u,v)$ means 2D feature extracted from images. $x$, $y$, $z$ mean coordinate in 3D space, $u$, $v$ mean coordinates on 2D space. There are challenges in deploying attention operations of transformer architecture on some computing platforms, which prevents the practical application of these methods.

The depth-based methods obtain the 3D BEV feature by calculating the outer product of the 2D feature and the predicted depth. The specific process is shown in formula~\ref{eq2}.
\begin{equation} \label{eq2}
\begin{aligned}
    F_{bev}(x,y,z) = Pool\{F_{2D}(u,v) \otimes D(u,v)\}_{x,y,z}     
\end{aligned}
\end{equation}
where $F_{2D}(u,v)$ means 2D feature extracted from images and $D(u,v)$ means depth prediction from 2d feature. $\otimes$ means outer product and $Pool$ means voxel pooling operation. $x$, $y$, $z$ mean coordinate in 3D space, $u$, $v$ mean coordinates on 2D space. With the support of CUDA multi-threading, these methods greatly improve the inference speed on the GPU platform, but the computational speed bottlenecks will be encountered at larger resolutions and feature dimensions, and it is not very friendly to transfer to non-GPU platforms without inference library support.

We propose the Fast-Ray transformation method, based on ray projection, with the help of Look-Up-Table and Multi-View to One-Voxel operations, to achieve the extremely fast 2D-to-3D inference speed on the GPU platforms. In addition, due to the high efficiency of its scalar index, when running on CPU platforms, it still has superior speed performance than existing solutions. This makes it possible to transfer to more platforms. See Sec.\ref{sec:projection} for details.

\begin{figure*}[t]
    \begin{center}
    \includegraphics[width=0.98\linewidth]{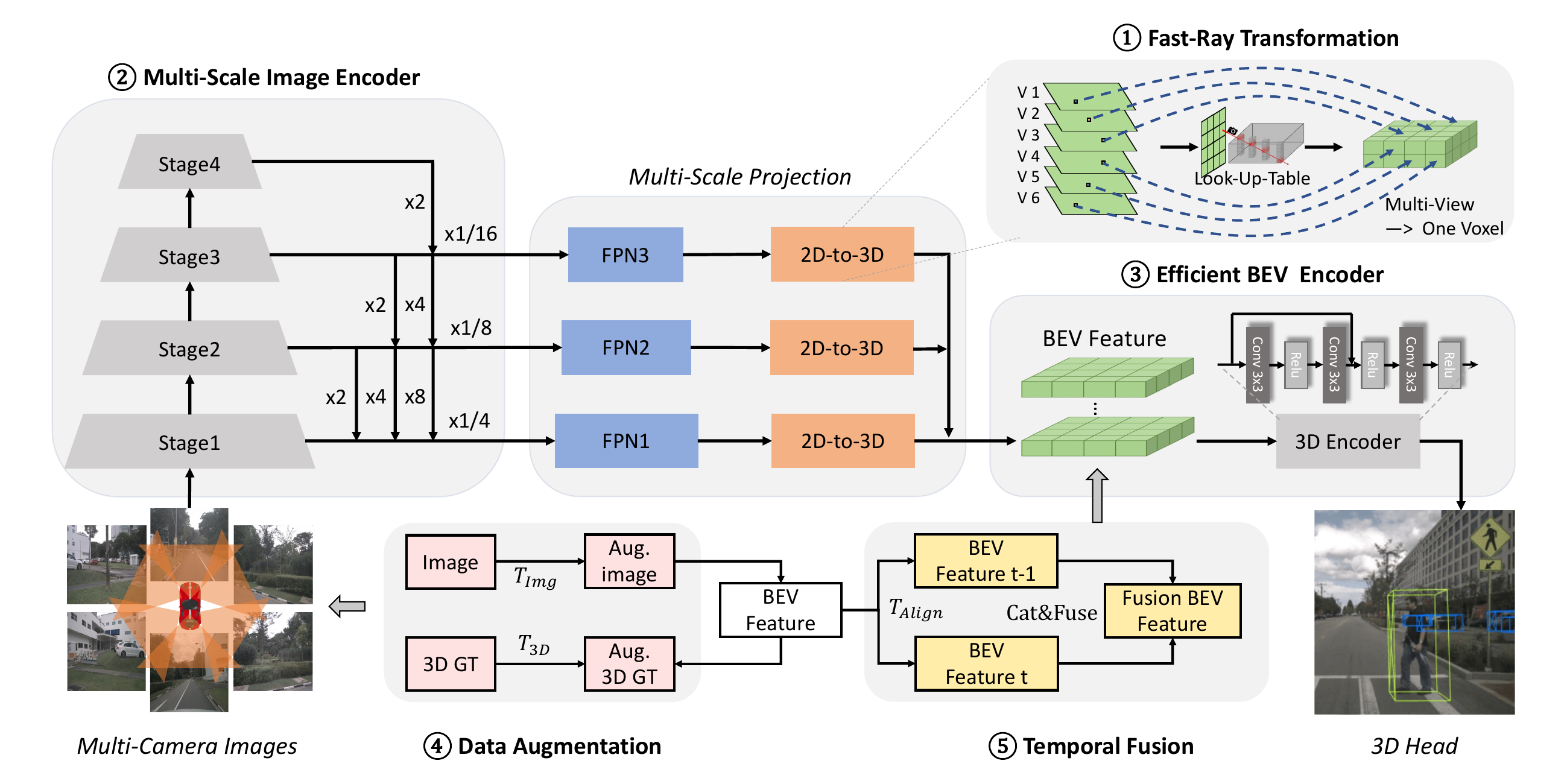}
    \end{center}
    \caption{Overview of \ours. It is consist of: \textcircled{1}\textbf{\textit{Fast-Ray Transformation}} with pre-computing the image-to-voxel index (Look-Up-Table) and letting all cameras project to the same dense voxel (Multi-View to One-Voxel) to speed up projection time, \textcircled{2} \textbf{\textit{Multi-Scale Image Encoder}} with Multi-Scale Projection to obtain multi-scale features, \textcircled{3} \textbf{\textit{Efficient BEV Encoder}} with efficient design to speed up inference time, \textcircled{4} \textbf{\textit{Data Augmentation}} on image and BEV domain to avoid over-fitting and achieve better performance, \textcircled{5} \textbf{\textit{Temporal Fusion}} module in BEV encoder stage to leverage multi-frame information.}
\label{fig:fast-bev_pipeline}
\end{figure*}

\subsection{Overview of \ours}\label{sec:overall}
M$^2$BEV~\cite{xie2022m} is one of the first works to solve the multi-camera multi-task perception with unified BEV representation, which has great application potential for on-vehicle platforms since it doesn't have expensive view transformer or depth representation. 
The input of M$^2$BEV is multi-camera RGB images, and the output is the predicted 3D bounding box and map segmentation results. It consists of a 2D image encoder that extracts image features from multi-camera images; an image-to-BEV (2D→3D) view transformation module that maps the 2D image feature into 3D BEV space; a 3D BEV encoder that processes the 3D features; and task-specific heads that perform perception tasks.
Inspired by its simplicity, we propose \ours with excellent speed and performance.
As denoted in the Fig.\ref{fig:fast-bev_pipeline}, \ours takes multi-camera images as input and predicts 3D bounding boxes (including velocity) as output. Its main framework can be divided into five key modules:

\textcircled{1} Fast-Ray Transformation. We find that the projection from image space to voxel space dominates the latency.
We propose Fast-Ray transformation, which projects multi-view 2D image features to 3D voxels along the camera rays, and two operations optimize the process for on-vehicle platforms. (1) We pre-compute the fixed projection indexes and store them as a static Look-Up-Table, which is super efficient during inference. 
(2) We let all the cameras project to the same voxel to avoid expensive voxel aggregation (Multi-View to One-Voxel). Our proposal is not like the improved view transformation schemes~\cite{li2022bevdepth,huang2021bevdet,liu2022bevfusion} based on Lift-Splat-Shoot. Even if they use cumbersome CUDA parallel computing, the inference speed is not optimal, let alone using the CPU to undertake the corresponding calculation process. The latency of Fast-BEV on the GPU is negligible, the inference speed on the CPU is far ahead of other solutions, and the deployment is very convenient. More details are in Sec~\ref{sec:projection}.

\textcircled{2} Multi-Scale Image Encoder.
Inspired by 2D detection tasks~\cite{lin2017feature,ren2015faster,lin2017focal} and CVT~\cite{zhou2022cross}, multi-scale design can bring performance improvement. We leverage the speed advantage brought by Fast-Ray transformation to design a multi-scale BEV perception paradigm, expecting to gain performance benefit from multi-scale information. In \ours, image encoder obtains multi-scale image features output from a unified single-scale image input by a 3-layer FPN output structure.
More details are in Section~\ref{sec:2d encoder}.

\textcircled{3} Efficient BEV Encoder.
Experiments have found that more blocks and larger resolutions in the 3D encoder resulted in a rapid increase in inference latency but did not significantly improve model performance. Along with "space-to-channel"(S2C), we only use one layer multi-scale concatenation fusion (MSCF) and multi-frame concatenation fusion (MFCF) module and less residual structure as BEV encoder, which greatly reduces the inference latency and has no damage to the accuracy. More details are in Section~\ref{sec:3d encoder}.

\textcircled{4} Data Augmentation. We empirically observe that severe over-fitting problem happened during the later training epochs in the basic framework. 
This is because no data augmentation is used in basic framework.
Motivated by recent work~\cite{huang2021bevdet,li2022bevdepth}, we add strong data augmentation on both image and BEV space, such as random flip, rotation \etc. More details are in Section~\ref{sec:aug}.

\textcircled{5} Temporal Fusion. In real autonomous driving scenarios, the input is temporally continuous and has tremendous complementary information across time. 
For instance, one pedestrian partially occluded at current frame might be fully visible in the past several frames.
Thus, we extend basic framework from spatial-only space to spatial-temporal space via introducing the temporal feature fusion module, similar to~\cite{huang2022bevdet4d,li2022bevformer}.
More specifically, we use current frame BEV features and stored history frame features as input and train \ours in an end-to-end manner. 
More details are in Section~\ref{sec:multiframe}.

We would like to clarify that \textcircled{4} and \textcircled{5} are inspired by the concurrent leading works~\cite{huang2022bevdet4d,li2022bevformer}, and we do not intend to regard these two parts as novel designs. 
These improvements make our proposed pipeline, \ours, become a competitive method while keeping its simplicity for on-vehicle platforms.

\subsection{Fast-Ray Transformation} \label{sec:projection}

\begin{figure}[t]
    \centering
    \includegraphics[width=1.0\linewidth]{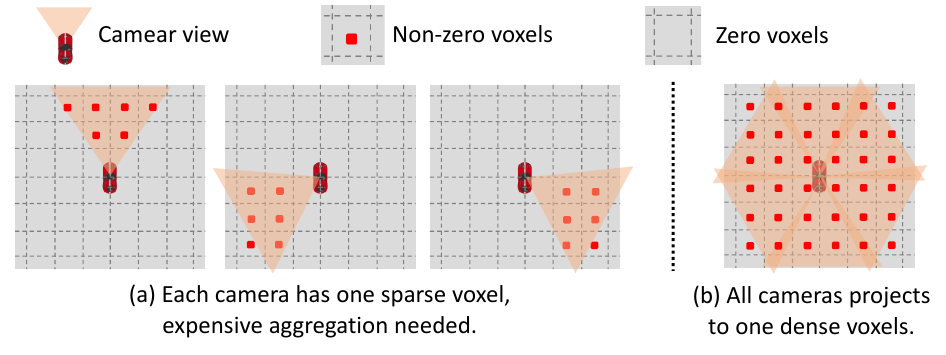}
    \caption{It is a bird’s-eye view of each discrete voxel filled. (a) In basic view transformation, each camera has one sparse voxel (only $\sim$17\% positions are non-zeros). An expensive aggregation operation is needed to combine the sparse voxels. (b) The proposed \ours let all cameras project to one dense voxel, avoiding the expensive voxel aggregation.
    }
\label{fig:accelerator}
\end{figure}

\textbf{Basic View Transformation.}
View transformation is the critical component to transform features from 2D image space to 3D BEV space, which typically takes much time in the whole pipeline. 
We follow \cite{xie2022m,murez2020atlas} to assume the depth distribution is uniform along the ray. The advantage is that once we get the intrinsic/extrinsic parameters of cameras, we can easily know the 2D-to-3D projection.
Since no learnable parameters are used here, we can easily compute the corresponding matrix between points in the image feature maps and the BEV feature maps. 
Based on this assumption, we further accelerate the process from two perspectives: pre-compute projection index (Look-Up-Table) and dense voxel feature generation (Multi-View to One-Voxel).

\begin{algorithm}
\caption{Build the Look-Up-Table}
\label{alg:code1}
\begin{algorithmic}[1]
\STATE \textbf{Inputs: } The Camera Parameter Matrix $projection$;  
\STATE \textbf{Output: } The Look-Up-Table $LUT$
\STATE

\FOR{\texttt{$offset=0,1,2,\cdots,volume\_size$}}

\STATE $LUT[offset] = (-1, -1, -1);$

\FOR{\texttt{$img=0,1,2,\cdots,img\_size$}}

\STATE Calculate the 2D pixel coordinates $(x, y, z)$ corresponding to the 3D coordinates $offset$ by $projection$.

\IF{$0 \leq x < w$ and $0 \leq y < h$ and $z \ge 0$}

\STATE $LUT[offset] = (img, x, y);$
\STATE $break;$
\ENDIF

\ENDFOR
\ENDFOR
\end{algorithmic}
\end{algorithm}

\textbf{Look-Up-Table.}
The projection index is the mapping index from 2D image space to 3D voxel space. Given the fact that camera positions and their intrinsic/extrinsic parameters are fixed when the perception system is built, and our method does not rely on the data-dependent depth prediction nor transformer, the projection index will be the \emph{same} for every input.
Thus, we don't need to compute the \emph{same} index for each iteration just pre-compute the fixed projection index and store it.
We propose to pre-compute the fixed projection indexes and store them as a static Look-Up-Table.
During inference, we can get the projection index via querying the Look-Up-Table, which is a very low-cost operation on edge devices.
Moreover, if we extend from the single frame to multiple frames, we can also easily pre-compute the intrinsic and extrinsic parameters and pre-align them to the current frame. 
As Shown in Alg.\ref{alg:code1}, we build the Look-Up-Table $LUT$ with the same dimension as the output 3D voxel space by the camera parameter matrix $projection$. We iterate over each voxel cell and calculate the 2D pixel coordinates corresponding to the 3D coordinates by $projection$. If the obtained 2D pixel coordinate addresses are legal, we fill them into the $LUT$ to establish a data-independent index mapping.

\begin{algorithm}
\caption{View Transformation with Look-Up-Table and Multi-View to One-Voxel}
\label{alg:code2}
\begin{algorithmic}[1]
\STATE \textbf{Inputs: } The Multi-view 2D Image Features $features$; The Fix Pre-compute Look-Up-Table $LUT$;
\STATE \textbf{Output: } The 3D Voxel Feature $volume$

\STATE

\FOR{\texttt{$offset=0,1,2,\cdots,volume\_size$}}

\STATE $img, x, y  = LUT[offset];$

\IF{\texttt{img >= 0}}

\STATE $volume[offset] = features[img][x][y];$

\ENDIF

\ENDFOR

\end{algorithmic}

\end{algorithm}

\textbf{Multi-View to One-Voxel.}
Basic view transformation uses the naive voxel aggregate operation, which stores a discrete voxel feature for each camera view and then aggregate them to generate the final voxel feature. As shown in Fig.\ref{fig:accelerator}(a), it is a bird's-eye view of each discrete voxel filled. 
Because each camera only has limited view angle, each voxel feature is very sparse, \eg, only about 17\% positions are non-zeros.
We identify the aggregation of these voxel features is very expensive due to their huge size.
We propose to generate a dense voxel feature to avoid the expensive voxel aggregation. 
Specifically, we let image features from all camera views project to the same voxel feature, leading to one dense voxel at the end, named Multi-View to One-Voxel. As shown in Fig.\ref{fig:accelerator}(b), it is a bird's-eye view of dense voxel filled. 
The Fast-Ray transformation algorithm as shown in Alg.\ref{alg:code2}, which transfer the input multi-view 2D image features into one-voxel 3D space, where each voxel cell is filled with the corresponding 2D image features by the pre-computed $LUT$. For the case of multiple views with overlapping areas, we directly adopt the first encountered view to improve the speed of table building.
Combining Look-Up-Table and Multi-View to One-Voxel acceleration designs, the view transformation operation has an extremely fast projection speed.

\subsection{Multi-Scale Image Encoder} \label{sec:2d encoder}
\begin{figure}[t]
    \centering
    \includegraphics[width=1.02\linewidth]{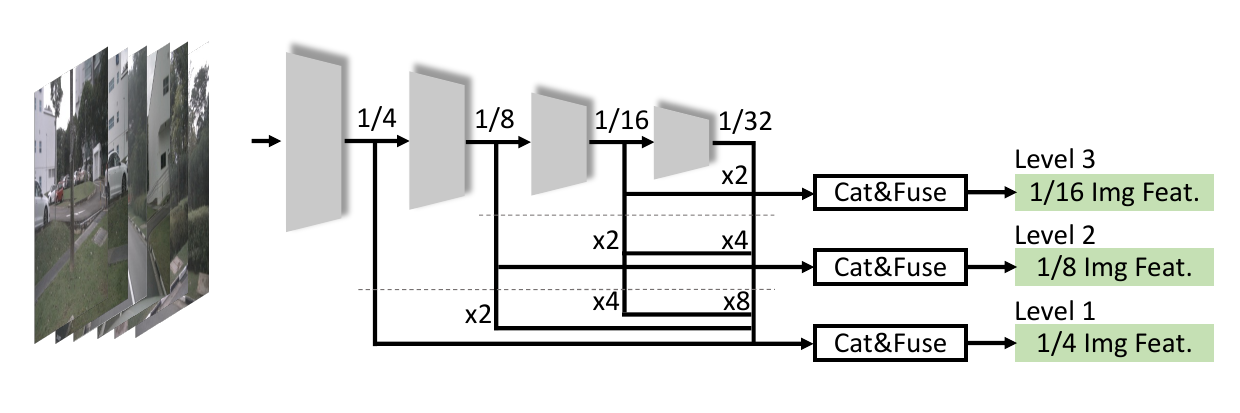}
    \caption{Multi-scale image encoder extracts multi-level features from multi-view images. $N$ images $\in R^{H \times W \times 3}$ as input and $F_{1/4}, F_{1/8}, F_{1/16}$ 3-level features as output.
    }
\label{fig:2d_encoder}
\end{figure}
The extremely fast Fast-Ray transformation allows backbone to leverage performance potential of multi-scale operation. As shown in Fig.\ref{fig:2d_encoder}, $N$ images $\in R^{H \times W \times 3}$ are input to image encoder network like ResNet-50\cite{he2016deep} to obtain 4-stage features $F_1, F_2, F_3, F_4$ with shape $\frac{H}{2^{i+1}}  \times \frac{W}{2^{i+1}} \times C$. Then we use a 3-layer multi-scale FPN structure in the image encoder output part. 
Each layer of FPN fuses the features of the same size after upsampling in the following layers through $1\times1$ convolution, and finally obtains $F_{1/4}, F_{1/8}, F_{1/16}$ 3-level comprehensive image features as output.

After extracting multi-scale image features from multi-view images, we project 3-level features $F = \{R^{N\times \frac{W}{i}\times \frac{H}{i}\times C}|i\in[4,8,16]\}$ by Fast-Ray transformation based multi-scale projection to obtain multi-scale BEV features $V = \{R^{X_i\times Y_i\times Z\times C}|X_i,Y_i\in[200,150,100]\}$.

\subsection{Efficient BEV Encoder} \label{sec:3d encoder}
BEV features are 4D tensor, and temporal fusion will stack the features, which will make the BEV encoder have a large amount of calculation. Three dimension reduction operations are used to speed up the encoder, which are the "space-to-channel" (S2C)~\cite{xie2022m} operator, the multi-scale concatenation fusion (MSCF) operator, and the multi-frame concat fusion (MFCF) operator, respectively. The S2C operator transforms a 4D voxel tensor $V \in R^{X\times Y\times Z\times C}$ to a 3D BEV tensor $V \in R^{X\times Y\times (ZC)}$, thus avoiding the usage of memory expensive 3D convolutions. Before the MFCF operator, it is worth noting that the BEV features obtained through multi-scale projection are different scales. We will first upsample the multi-scale BEV features in the $X$ and $Y$ dimensions to the same size, such as $200\times200$. The MSCF\&MFCF operators concat multi-scale multi-frame features in the channel dimension and fuses them from a higher parameter amount  to a lower parameter amount, $Fuse(V_s|V_s \in R^{X\times Y\times (ZCF_{s}T)})$ $\implies$ $V|V \in R^{X\times Y\times C_{MSCF\&MFCF}}$, $s\in 3-level$ indicates multi-scale BEV features with three strides, thereby speeding up the computation time of the BEV encoder. In addition, it is found through experiments that the number of blocks of the BEV encoder and the size of the 3D voxel resolution have relatively little impact on performance, but occupy a large speed consumption, so fewer blocks and smaller voxel resolutions are also more crucial.

\subsection{Data Augmentation} \label{sec:aug}
The benefits of data augmentation have formed a consensus in academia. In addition, 3D datasets, such as NuScenes\cite{caesar2020nuscenes}, KITTI\cite{Geiger2013IJRR}, are difficult and expensive to label, which results in a small number of samples in the dataset, so data augmentation can bring more significant performance gains. We add data augmentations in both image space and BEV space, mainly following BEVDet~\cite{huang2021bevdet}. 

\begin{figure}[t]
    \centering
    \includegraphics[width=1\linewidth]{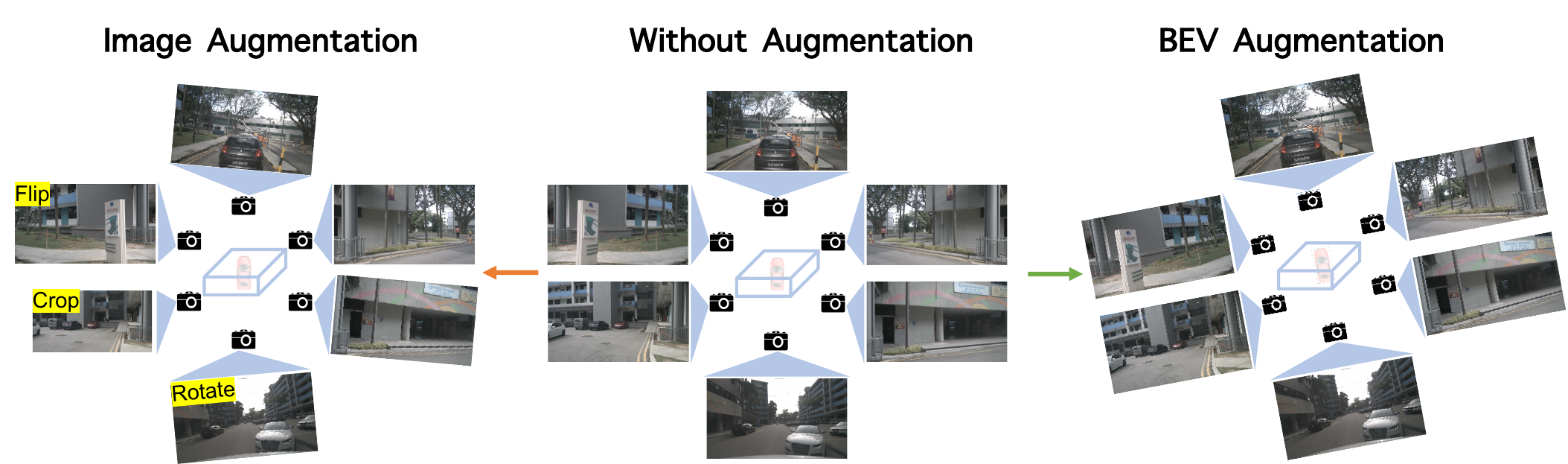}
    \caption{Examples of the data augmentation used in Fast-BEV. 
    The middle figure shows no data augmentation.
    The left figure shows the image augmentation and some augmentation types such as random flip, crop and rotate. 
    The right figure shows one type of BEV augmentation, random rotation.
    }
\label{fig:aug_fig}
\end{figure}

\textbf{Image Augmentation.} The data augmentation in 3D object detection is more challenging than that of 2D detection since the images in 3D scenarios have direct relationship with 3D camera coordinates. Thus, if we apply data augmentation on images, we also need to change the camera intrinsic matrix~\cite{huang2021bevdet}.
For the augmentation operations, we basically follow the common operations, \eg, flipping, cropping and rotation.
In the left part of Fig.\ref{fig:aug_fig}, we show some examples of image augmentations.

\textbf{BEV Augmentation.} Similar to image augmentation, similar operations can be applied to the BEV space, such as flipping, scaling and rotation. Note that the augmentation transformation should be applied on both the BEV feature map and the 3D ground-truth box to keep consistency. The BEV augmentation transformation can be controlled by modifying the camera extrinsic matrix accordingly. 
In the right part of Fig.\ref{fig:aug_fig}, we show the random rotation augmentation, a type of BEV augmentation.

\subsection{Temporal Fusion} \label{sec:multiframe}

\begin{figure}[h]
  \centering
  \includegraphics[width=0.4\textwidth]{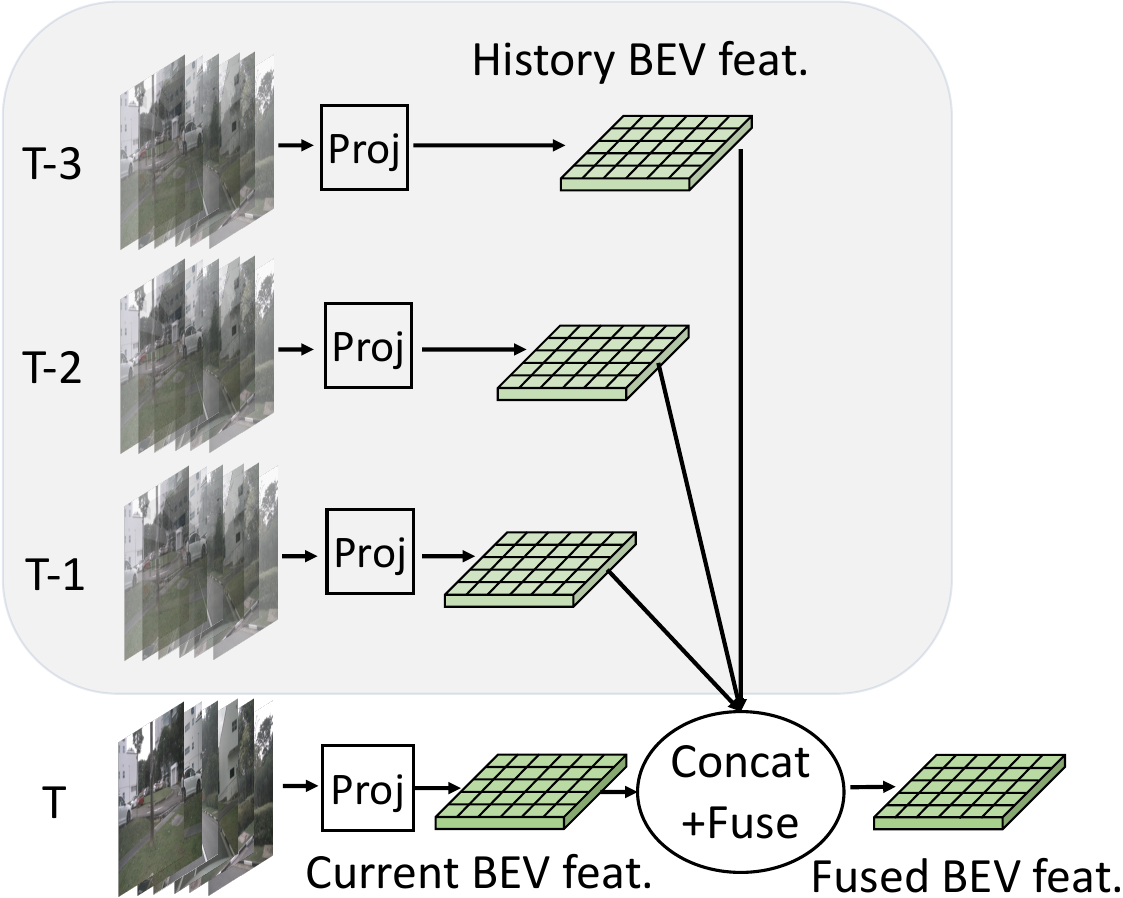}
  \caption{Illustration of the temporal multi-frame feature fusion module. The three history frames are first extracted features and projected to the respective BEV space, then aligned to the current frame with camera extrinsic and global coordinate. Finally, we directly concatenate these multi-frame BEV features in the channel dimension.}
  \label{fig:fig_fuse}
\end{figure}

Inspired by BEVDet4D~\cite{huang2022bevdet4d} and BEVFormer~\cite{li2022bevformer}, we also introduce the history frame into the current frame for temporal feature fusion. The features of the history frames are fused with the corresponding features of the current frame through the spatial alignment operation and the concatenation operation. Temporal fusion can be considered as feature augmentation at the frame level, and longer time sequences within a certain range can bring more performance gains. Specifically, we sample the current frame with three history key frames; each key frame has a 0.5s interval.
We adopted the multi-frame feature alignment method from BEVDet4D. 
As shown in Fig.~\ref{fig:fig_fuse}, after we got four aligned BEV features, we directly concatenate them and feed them to the BEV encoder.
In the training phase, the history frame features are extracted online using the image encoder.
In the testing phase, the history frame feature can be saved offline and directly taken out for acceleration.

\textbf{Compared with BEVDet4D and BEVFormer.}
BEVDet4D only introduces one history frame, which we argue is insufficient to leverage history information. 
\ours uses three history frames, resulting in significant performance improvement. 
BEVFormer is slightly better than BEVDet4D by using two history frames. However, due to memory issues, in the training phase, the history feature is detached without gradient, which is not optimal. Moreover, BEVFormer uses an RNN-style to sequentially fuse features, which is inefficient. 
In contrast, all the frames in \ours is trained in an end-to-end manner, which is more training friendly with common GPUs.

\subsection{Detection Head}\label{sec:detection}
Referring to PointPillars~\cite{lang2019pointpillars}, we use three parallel $1\times1$ convolutions to perform 3D detection tasks on BEV features, which can predict category, box size, and direction
of each object.  The loss is the same as PointPillars:

\begin{equation}
     \mathcal{L}_{det} = \frac{1}{N_{pos}} (\beta_{cls}\mathcal{L}_{cls} + \beta_{loc}\mathcal{L}_{loc} + \beta_{dir}\mathcal{L}_{dir})
    \label{eqn:lossdet}
\end{equation}

where $N_{pos}$ is the number of positive anchors,  ${L}_{loc}$ is Focal Loss, ${L}_{loc}$ is Smooth-L1 loss, and ${L}_{dir}$ is binary cross-entropy loss.

\begin{figure*}[t]
    \centering
    \includegraphics[width=1\linewidth]{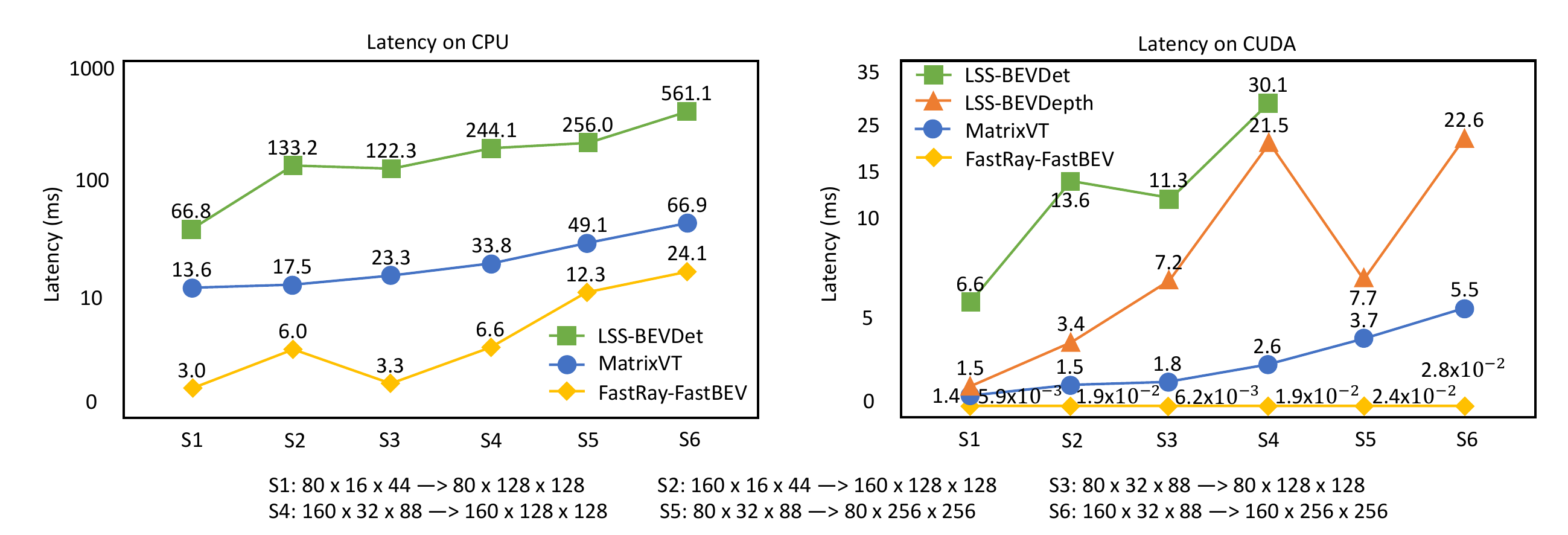}
    \caption{Following MatrixVT\cite{zhou2022matrixvt}, compare the latency of \ours and several mainstream methods, BEVDet\cite{huang2021bevdet}, BEVDepth\cite{li2022bevdepth}, MatrixVT\cite{zhou2022matrixvt}, in the view transformation part. The left figure shows the CPU latency, and the right figure shows the CUDA latency. S1$ \sim $S6 represent different image feature sizes ($C\times H_{Img}\times W_{Img}$) to different BEV feature sizes ($C\times H_{BEV}\times W_{BEV}$). The baseline results other than Fast-BEV are all from MatrixVT, and the lack of latency measurement of LSS-BEVDepth on the CPU is due to the fact that LS-BEVDepth cannot be used without a CUDA platform.}
    \label{fig:efficient_fig}
\end{figure*}

\begin{table*}[h]
\centering
\caption{End-to-end latency comparison of \ours, BEVDet4D\cite{huang2022bevdet4d}, and BEVDepth\cite{li2022bevdepth} schemes under comparable performance. The upper part of table is the detailed setting of the three schemes, including the specific configuration of each component. The lower part of table is a comparison of the latency of each part and the overall latency of the three schemes under comparable performance. The 2D-to-3D part includes CPU and CUDA two platform latency. }
\label{table:efficient_setting}
\resizebox{1\textwidth}{!}{
\begin{tabular}{c|c|c|c|c|c|c}
 
\noalign{\smallskip}\hline\noalign{\smallskip}
Method & Image Res.& Image Encoder& 2D-to-3D & BEV Res. &BEV Encoder & Detection Head \\
\noalign{\smallskip}\hline\noalign{\smallskip}

BEVDet4D&$256\times704$&R50\&LSS-FPN&LSS-BEVDet&$128\times128$&ResBlock\&FPN-LSS&CenterPoint\\
\noalign{\smallskip}\hline\noalign{\smallskip}
BEVDepth&$256\times704$&R50\&LSS-FPN&LSS-BEVDepth&$128\times128$&ResBlock\&FPN-LSS&CenterPoint \\
\noalign{\smallskip}\hline\noalign{\smallskip}
\ours&$256\times704$&R50\&FastBEV-FPN&FastRay&$200\times200\times4$&Efficient-ResBlock&PointPillar \\
\noalign{\smallskip}\hline\noalign{\smallskip}

\noalign{\smallskip}\hline\noalign{\smallskip}
Method & mAP &NDS &2D CUDA Latency(ms) &2D-to-3D CUDA/CPU Latency(ms) &3D CUDA Latency(ms) &Overall Latency(ms) \\
\noalign{\smallskip}\hline\noalign{\smallskip}

BEVDet4D&0.322&0.457&10.8&$6.6\times2$ / $66.8\times2$&9.1&33.1 (30.2 fps)/153.5 (6.5 fps)\\
\noalign{\smallskip}\hline\noalign{\smallskip}
BEVDepth&0.351&0.475&12.1&$1.5\times2$ / $66.8\times2$&9.1&24.2 (41.3 fps)/154.8 (6.5 fps) \\
\noalign{\smallskip}\hline\noalign{\smallskip}

\ours&0.334&0.473&11.3&$0.006\times4$ / $7.0\times4$ &7.7&19.0 (52.6 fps)/47.0 (21.3 fps) \\
\noalign{\smallskip}\hline\noalign{\smallskip}

\end{tabular}
}
\end{table*}

%% file: Sections/experiments.tex
\section{Experiments}
\subsection{Setup}

\textbf{Dataset Description}
We evaluate our \ours on nuScenes dataset~\cite{caesar2020nuscenes}, which contains 1000 autonomous driving scenes with 20 seconds per scene.
The dataset is split into 850 scenes for training/validation and the rest 150 for testing.  
While nuScenes dataset provides data from different sensors, we only use the camera data. 
The cameras have six views: $\tt front\_left$, $\tt front$, $\tt front\_right$, $ \tt back\_left$, $\tt back$, $\tt back\_right$. 

\textbf{Evaluation metrics.} 
To comprehensively evaluate the detection task, we use the standard evaluation metrics of mean Average Precision (mAP), and nuScenes detection score (NDS) for 3D object detection evaluation.
In addition, in order to calculate the precision of the corresponding aspects (\textit{e.g.}, translation, scale, orientation, velocity, and attribute), we use the mean Average Translation Error (mATE), mean Average Scale Error (mASE), mean Average Orientation Error(mAOE), mean Average Velocity Error(mAVE), and mean Average Attribute Error(mAAE) as the metrics.

\textbf{Implementation Details.}
For training, we use AdamW optimizer with learning rate $1e^{-3}$ and the weight decay is set to $1e^{-2}$.
``PolyLR'' scheduler is adopted to gradually decrease the learning rate. 
We also use ``warmup'' strategy for the first 1000 iterations.
For data augmentation hyper-parameters, we basically follow BEVDet~\cite{huang2021bevdet}.
In addition, we use a dynamic box assignment to assign anchors with ground-truth and an empirical loss weight, which are learning from M$^2$BEV~\cite{xie2022m}. Specifically, $\beta_{cls}=1.0$, $\beta_{loc}=0.8$,  $\beta_{dir}=0.8$, and 3D boxes are defined by $(x,y,z,w,h,l, \theta, v_{x}, v_{y})$ with loss weight $[1, 1, 1, 1, 1, 1, 1, 0.2, 0.2]$ for each item.
Without specific notifications, all models are trained on 32 A100 GPUs for 48 epochs.
For comparison with state-of-the-art, we train 20 epochs with CBGS~\cite{zhu2019class,mmdet3d2020}, and the voxel resolution is $200\times200\times6$ with four frames. 
We mainly use ResNet-50 as the backbone for the ablation study to verify the ideas quickly. 
For on-vehicle inference speed test, we test with one sample per batch which contains 6 view images.

\subsection{Speed Comparison with Popular Methods}
\textbf{Comparison on View Transformation Latency.}
View transformation affects the design of the BEV perception solution, the deployability and speed of the overall model on different platforms. We compare proposed Fast-Ray transformation with other popular LSS-based view transformation methods, such as LSS-DEVDet\cite{huang2021bevdet}, LSS-BEVDepth\cite{li2022bevdepth}, and MatrixVT\cite{zhou2022matrixvt}, on both CPU and GPU platforms. For a fair comparison, We follow the same experimental setting of MatrixVT for speed comparison, which takes into account the three main factors that affect the view transformation latency: the channel size, the image resolution and the voxel size. Specifically, the LSS-BEVDet is the accelerated view transformation used in BEVDet; the LSS-BEVDepth is the CUDA multi-threaded view transformation in BEVDepth, and it is not available on CPU and other platforms\cite{zhou2022matrixvt}; the LSS-MatrixVT is the view transformation proposed in MatrixVT. In order to comprehensively compare several different solutions, we compare 6 different settings from S1 to S6 on the two platforms of CPU and GPU, where $S_i$ indicate projecting different image feature sizes to different BEV feature sizes. All latency measurements were performed on 2080Ti device.

As shown in Fig.\ref{fig:efficient_fig}, we show the view transformation latency comparison results of our Fast-Ray transformation and other methods. The left figure of Fig.\ref{fig:efficient_fig} shows the CPU latency, we can know that Fast-Ray transformation has a great speed advantage over LSS-BEVDet and LSS-MatrixVT, whether it is the smaller-dimensional S1 setting or the larger-dimensional S6 setting. Specifically, the S1 setting Fast-Ray is 22 times faster than LSS-BEVDet and 4 times faster than LSS-MatrixVT; the S6 setting Fast-Ray is 23 times faster than LSS-BEVDet and 3 times faster than LSS-MatrixVT; The rest of the settings have a similar proportional speed advantage. For some non-GPU computing platforms without CUDA inference acceleration library, such as hardware with DSP as the computing unit, CPU will be the first choice for this part of the operation if no new operator is developed to support view transformation operation. In this case, Fast-Ray shows considerable application potential. 

The right figure of Fig.\ref{fig:efficient_fig} shows the CUDA latency, we can see that from S1 to S6 settings, the latency of Fast-Ray is negligible compared with LSS-BEVDet, LSS-BEVDepth and LSS-MatrixVT. Fast-Ray compresses the latency of view transformation to the extreme, especially when the feature size is relatively large, and the speed advantage is more significant than other methods. For example, the S3 setting is three orders of magnitude faster than LSS-BEVDet and BEVDepth, and two orders of magnitude faster than MatrixVT.
\begin{table*}[t]
\centering
\caption{Comparison on the nuScenes \emph{val} set. ``L'' denotes LiDAR, ``C'' denotes camera and ``D'' denotes Depth/LiDAR supervision. 
MS indicates multi-scale in image and BEV encoders.
``\P'' indicates our method with MS, scale NMS and test-time augmentation. ``T'' indicates the use of temporal, and ``F'' is the opposite.
}
\resizebox{\textwidth}{!}{
\begin{tabular}{l|c|c|c|cccccc|c}
\hline\noalign{\smallskip}
Methods & Image Res. &  Modality &Temporal& mAP$\uparrow$ & mATE$\downarrow$ &mASE$\downarrow$ &mAOE$\downarrow$ &mAVE$\downarrow$ &mAAE$\downarrow$ & NDS$\uparrow$\\
\noalign{\smallskip}
\hline
\noalign{\smallskip}
CenterPoint-Voxel~\cite{yin2021center}  &         -      & L        & &0.564 &   -   &    -   &     -    &     -  &    -   & 0.648  \\
CenterPoint-Pillar~\cite{yin2021center} &       -        & L        & &0.503 &   -    &    -   &     -    &  -     &     -  & 0.602  \\ 

\noalign{\smallskip}
\hline
\noalign{\smallskip}
\noalign{\smallskip}
\hline
\noalign{\smallskip}
FCOS3D~\cite{wang2021fcos3d}             & 900$\times$1600      & C        &F& 0.295 & 0.806 & 0.268 & 0.511   & 1.315 & 0.170 & 0.372  \\
BEVDet-R50~\cite{huang2021bevdet}         & 256$\times$704       & C        &F& 0.286 & 0.724 & 0.278 & 0.590   & 0.873 & 0.247 & 0.372 \\
PETR-R50~\cite{liu2022petr}           & 384$\times$1056      & C        &F &0.313 & 0.768 & 0.278 & 0.564   & 0.923 & 0.225 & 0.381  \\
PETR-Tiny~\cite{liu2022petr} & 512$\times$1408      & C        & F&0.361 & 0.732 & 0.273 & 0.497   & 0.808 & 0.185 & 0.431  \\
BEVDet4D-Tiny(v1)~\cite{huang2022bevdet4d}      & 256$\times$704       & C     &  T & 0.323 & 0.674 & 0.272 & 0.503   & 0.429 & 0.208 & 0.453  \\
BEVDet4D-Tiny(v3)~\cite{huang2022bevdet4d}      & 256$\times$704       & C      &T  & 0.338 & 0.672 & 0.274 & 0.460 &0.337 &0.185 &0.476  \\
BEVDepth-R50~\cite{li2022bevdepth}        & 256$\times$704       & C\&D       & T& 0.351 & 0.639 & 0.267 & 0.479   & 0.428 & 0.198 & 0.475  \\
\noalign{\smallskip}
\hline
\noalign{\smallskip}
\textbf{Fast-BEV(R50)}  & 256$\times$704& C &T&0.334 & 0.665 &0.285 & 0.393   & 0.388 & 0.210 &0.473\\
\textbf{Fast-BEV(R50)MS}  & 256$\times$704& C &T& 0.343 & 0.647 & 0.282 & 0.360 & 0.342 & 0.225 & 0.485 \\
\textbf{Fast-BEV(R50)\P}  & 256$\times$704& C &T& 0.354 & 0.656 & 0.281 & 0.384 & 0.361 & 0.217 & 0.487 \\
\hline
FCOS3D-R101$\dagger$~\cite{wang2021fcos3d}   & 900$\times$1600& C        &F& 0.321 & 0.754 & 0.260 & 0.486   & 1.331 & 0.158 & 0.395  \\
DETR3D-R101$\dagger$~\cite{wang2022detr3d}    & 900$\times$1600 & C       & F& 0.347 & 0.765 & 0.267 & 0.392   & 0.876 & 0.211 & 0.422  \\
Ego3RT-V2-99$\dagger$~\cite{lu2022learning}   & 900$\times$1600 & C    &F& 0.478 & 0.582 & 0.272 & 0.316   & 0.683 & 0.202 & 0.534  \\
M$^2$BEV-X101$\dagger$~\cite{xie2022m}  & 900$\times$1600 & C    &F& 0.417 & 0.647 & 0.275 & 0.377   & 0.834 & 0.245 & 0.470  \\
PolarFormer-T-R101$\dagger$~\cite{jiang2022polarformer}   & 900$\times$1600  & C   & T& 0.432 & 0.648 & 0.270 & 0.348   & 0.409 & 0.201 & 0.528  \\
PETRv2-R101$\dagger$~\cite{liu2022petrv2}  & 640$\times$1600     & C   & T& 0.421 &0.681 &0.267 &0.357 &0.377 &0.186 &0.524  \\
DETR3D~\cite{wang2022detr3d}             & 900$\times$1600      & C       & F& 0.303 & 0.860 & 0.278 & 0.437   & 0.967 & 0.235 & 0.374  \\
BEVDet-Base~\cite{huang2021bevdet}       & 512$\times$1408      & C       &F & 0.397 &0.595 &0.257 &0.355 &0.818 &0.188 &0.477\\
PETR-R101 ~\cite{liu2022petr}         & 512$\times$1408      & C       &F & 0.357 & 0.710 & 0.270 & 0.490   & 0.885 & 0.224 & 0.421  \\
BEVDet4D-Base(v1) ~\cite{huang2022bevdet4d}     & 640$\times$1600      & C       &T & 0.396 & 0.619 & 0.260 & 0.361   & 0.399 & 0.189 & 0.515  \\
BEVDet4D-Base(v3) ~\cite{huang2022bevdet4d}     & 640$\times$1600      & C       &T & 0.426 &0.560 &0.254 &0.317 &0.289 &0.186 &0.552  \\
BEVFormer-R101~\cite{li2022bevformer}        &        900$\times$1600       & C      &  T& 0.416 & 0.673 & 0.274 & 0.372   & 0.394 & 0.198 & 0.517  \\
BEVDepth-R101 ~\cite{li2022bevdepth}      & 512$\times$1408      & C\&D      &T  & 0.412 & 0.565 & 0.266 & 0.358   & 0.331 & 0.190 & 0.535  \\
\noalign{\smallskip}
\hline
\noalign{\smallskip}
\textbf{Fast-BEV(R101)} & 900$\times$1600& C & T& 0.402 & 0.582 &0.278 &0.304   & 0.328 & 0.209 & 0.531\\
\textbf{Fast-BEV(R101)\P} & 900$\times$1600& C& T & 0.413 & 0.584 & 0.279 & 0.311 & 0.329 & 0.206 & 0.535 \\
\hline
\end{tabular}}
\label{tab:val_performance}
\end{table*}
\setlength{\tabcolsep}{1.4pt}

\textbf{Comparison on End-to-End Deployment Latency.} Based on the implementation of view transformation, we compare the latency of \ours, BEVDet4D and BEVDepth schemes after deploying on 2080Ti platform with CUDA11.1-TRT7.2.1-FP32. As shown in Table.\ref{table:efficient_setting}, the upper part of table is the detailed setting of the three schemes, including the specific configuration of each component, consisting of \textit{Image Resolution, Image Encoder, 2D-to-3D,  BEV Resolution, BEV Encoder, and Detection Head.} The lower part of table is a comparison of the latency of each part and the overall latency of the three schemes under comparable performance. Specifically, the 2D-to-3D part includes CPU and CUDA two platforms. It can be seen from the table that although the types of these components used in the three solutions may be different, they have basically similar performance, and it is worth noting that \ours uses 4-frame fusion, and the other two use 2-frame fusion. Under such a setting with comparable performance, we compared their latency after deployment, from the last column in the lower part of Table.\ref{table:efficient_setting}, we can see that our \ours has a smaller latency on CUDA, and a more obvious latency advantage when 2D-to-3D on CPU. Specifically, \ours can reach 52.6 FPS on CUDA, which is better than 30.2 FPS of BEVDet4D and 41.3 FPS of BEVDepth. When 2D-to-3D on CPU, \ours can reach 21.3 FPS, which is 3 times faster than the other two methods. In addition, as can be seen from the two columns of 2D-to-3D view transformation and 3D encoder, the latency advantage of \ours mainly comes from these two parts, which is also in line with the motivation of our design of Fast-Ray transformation and efficient BEV encoder.

\subsection{Performance Compared with Other Methods}
We comprehensively compare the proposed \ours with the lidar-based, camera-based, and camera-depth-based methods, such as PointPillar\cite{lang2019pointpillars}, FCOS3D~\cite{wang2021fcos3d}, PETR~\cite{liu2022petr}, BEVDet~\cite{huang2021bevdet}, DETR3d~\cite{wang2022detr3d}, BEVDet4D~\cite{huang2022bevdet4d}, BEVFormer~\cite{li2022bevformer}, BEVDepth~\cite{li2022bevdepth} and so on, on nuScenes val set.

As shown in Table.\ref{tab:val_performance}, \ours shows superior performance in mAP and NDS compare with existing advanced methods. 
For example, with ResNet-50\cite{he2016deep} as the image encoder, \ours achieves naive 0.334 mAP and 0.473 NDS, significantly outperforms the Camera-Based methods with similar level image encoder and image resolution, such as BEVDet4D-Tiny~\cite{huang2022bevdet4d,liu2021swin} (0.323 mAP and 0.453 NDS). The model also outperforms other methods with a larger input resolution such as PETR-R50~\cite{liu2022petr} (0.313 mAP and 0.381 NDS). When we use MS, scale NMS and test-time augmentation configurations, \ours ResNet-50 has stronger 0.354 mAP and 0.487 NDS, which can exceeds the Camera-Depth-Based BEVDepth\cite{li2022bevdepth}(0.351 mAP and 0.475 NDS). 

Moreover, with the larger image encoder ResNet-101 and higher image resolution, \ours establishes a better competitive 0.402 mAP and 0.531 NDS. When we use MS, scale NMS and test-time augmentation configurations, \ours ResNet-101 has stronger 0.413 mAP and 0.535 NDS, which exceeds BEVDet4D-Base~\cite{huang2022bevdet4d} 0.515 NDS and BEVFormer-R101~\cite{li2022bevformer} 0.517 NDS, and even ties BEVDepth-R101\cite{li2022bevdepth} 0.535 NDS.

\ours mainly emphasizes efficiency and deployment. Although it does not use any depth information during the view transformation process, nor any explicit and implicit depth information supervision, it has competitive performance compared to existing methods, even better than BEVDepth\cite{li2022bevdepth} with depth supervision information. This is mainly due to the design of our overall model architecture around Fast-Ray. The design of our image encoder, BEV encoder and detection head are different from BEVDepth. In addition, our pre-training strategy and multi-scale feature strategy are also different from BEVdepth. These have contributed to the competitive performance of our \ours. See Sec.\ref{sec:ablation} for details on the impact of each component on performance.
\vspace{-2em}

\subsection{Detailed Ablation Analysis}\label{sec:ablation}
\vspace{-1em}
Unless explicitly stated, Fast-BEV related analysis experiments for the detection task in this section use the ResNet-50 backbone and 2 frames to train 48 epochs without multi-scale feature. And the image resolution and voxel resolution are $256\times704$ and $200\times200\times6$, respectively.
\begin{minipage}{\textwidth}
\begin{minipage}[t]{0.31\columnwidth}
\makeatletter\def\@captype{table}
\caption{Ablation study of different augmentations with single frame.}
\centering
\setlength{\tabcolsep}{2mm}{
\begin{tabular}{c|c|c}
    \noalign{\smallskip}\hline\noalign{\smallskip}
    Method  & mAP & NDS\\
    \noalign{\smallskip}\hline\noalign{\smallskip}
    Baseline (M$^2$BEV)&  0.247 & 0.329\\
    +ImgAug &  0.285 & 0.357 \\
    +BEVAug &  0.265 & 0.352 \\
    +ImgBEVAug &0.293 & 0.373 \\
    \noalign{\smallskip}\hline\noalign{\smallskip}
\end{tabular}}
\label{table:augmentation}
\end{minipage}
\hspace{2mm}
\begin{minipage}[t]{0.31\columnwidth}
\makeatletter\def\@captype{table}
\caption{Ablation study of sequential feature fusion from single frame to 4 frames. }
\centering
\setlength{\tabcolsep}{2mm}{
\begin{tabular}{c|c|c}
    \noalign{\smallskip}\hline\noalign{\smallskip}
    Method & mAP & NDS\\
    \noalign{\smallskip}\hline\noalign{\smallskip}
    1F &  0.293 & 0.373 \\
    2F & 0.321 & 0.451 \\
    4F & 0.323 & 0.466 \\
    \noalign{\smallskip}\hline\noalign{\smallskip}
\end{tabular}}
\label{table:sequantial}
\end{minipage}
\hspace{2mm}
\begin{minipage}[t]{0.31\columnwidth}
\makeatletter\def\@captype{table}
\caption{Ablation study of multi-scale (MS) image and BEV features.}
\centering
\setlength{\tabcolsep}{2mm}{
\begin{tabular}{c|c|c}
    \noalign{\smallskip}\hline\noalign{\smallskip}
     Method  & mAP & NDS\\
    \noalign{\smallskip}\hline\noalign{\smallskip}
    Baseline (Fast-BEV 2F)& 0.321 & 0.451 \\
    Image MS& 0.320 & 0.451 \\
    BEV MS& 0.316 & 0.446 \\
    Image \& BEV MS& 0.324 & 0.455 \\

    \noalign{\smallskip}\hline\noalign{\smallskip}
\end{tabular}}
\label{table:multi-scale}
\end{minipage}
\end{minipage}

\begin{minipage}{\textwidth}
\begin{minipage}[t]{0.45\columnwidth}
\makeatletter\def\@captype{table}
\caption{Ablation study of image resolution.}
\centering
\setlength{\tabcolsep}{4mm}{
\begin{tabular}{l|c|c}
    \noalign{\smallskip}\hline\noalign{\smallskip}
    Image Res.  & mAP & NDS\\
    \noalign{\smallskip}\hline\noalign{\smallskip}
256$\times$448  &0.280 & 0.419 \\
256$\times$704  &0.321 & 0.451 \\
464$\times$800  &0.342 & 0.466 \\
544$\times$960  &0.345 & 0.472 \\
704$\times$1208 &0.358 & 0.478 \\
832$\times$1440 &0.368 & 0.491 \\
928$\times$1600 &0.369 & 0.488 \\
    \noalign{\smallskip}\hline\noalign{\smallskip}
\end{tabular}}
\label{table:s_input_size}
\end{minipage}
\hspace{2mm}
\begin{minipage}[t]{0.45\columnwidth}
\makeatletter\def\@captype{table}
\caption{Ablation study of voxel resolution. The resolution of the image is 384$\times$1056.}
\centering
\setlength{\tabcolsep}{4mm}{
\begin{tabular}{l|c|c}
    \noalign{\smallskip}\hline\noalign{\smallskip}
    Voxel Res.  & mAP & NDS\\
    \noalign{\smallskip}\hline\noalign{\smallskip}
    200$\times$200$\times$4& 0.345 & 0.472 \\
    200$\times$200$\times$6& 0.352 & 0.476 \\
    200$\times$200$\times$12& 0.350 & 0.474 \\
    300$\times$300$\times$6& 0.347 & 0.471 \\
    400$\times$400$\times$6& 0.337 & 0.467 \\
    400$\times$400$\times$12& 0.345 & 0.476 \\
    \noalign{\smallskip}\hline\noalign{\smallskip}
\end{tabular}}
\label{table:v_input_size}
\end{minipage}
\end{minipage}

\begin{minipage}{\textwidth}
\begin{minipage}[t]{0.45\columnwidth}
\makeatletter\def\@captype{table}
\caption{Ablation study of baseline and Fast-BEV with different epochs.}
\label{table:epoch}
\centering
\setlength{\tabcolsep}{4mm}{
\begin{tabular}{l|c|c|c}
    \noalign{\smallskip}\hline\noalign{\smallskip}
Method& Epochs & mAP & NDS\\
    \noalign{\smallskip}\hline\noalign{\smallskip}
Baseline &12 & 0.258 & 0.330\\
&24 & 0.257 & 0.338\\
&36 & 0.248 & 0.320\\
&48 & 0.247 & 0.329\\
    \noalign{\smallskip}\hline\noalign{\smallskip}
Fast-BEV&12&0.273 & 0.381 \\
&24&0.294 & 0.424 \\
&36&0.310 & 0.440 \\
&48&0.321 & 0.451 \\
    \noalign{\smallskip}\hline\noalign{\smallskip}
\end{tabular}}
\end{minipage}
\hspace{2mm}
\begin{minipage}[t]{0.45\columnwidth}
\makeatletter\def\@captype{table}
\caption{Ablation study of 2D and 3D encoders with 4 frames, 2D with 20 CBGS epochs and 3D with 48 epochs.}
\centering
\setlength{\tabcolsep}{4mm}{
\begin{tabular}{l|c|c|c}
        \noalign{\smallskip}\hline\noalign{\smallskip}
Encoder&Type& mAP & NDS\\
    \noalign{\smallskip}\hline\noalign{\smallskip}
2D&R18& 0.293 & 0.437 \\
&R34&0.331&0.467\\
&R50& 0.335 & 0.473 \\
&R101&0.345 & 0.482 \\
    \noalign{\smallskip}\hline\noalign{\smallskip}
3D &$1\times$Block   &0.305&0.432 \\
&2$\times$Block   &0.320&0.446 \\
&4$\times$Block   &0.315&0.448 \\
&6$\times$Block   &0.321&0.451 \\
    \noalign{\smallskip}\hline\noalign{\smallskip}
\end{tabular}}
\label{table:encoder}
\end{minipage}
\end{minipage}

\textbf{Resolution.} To investigate the effect of different resolutions of the input image and voxel, we perform an ablation study in Table.\ref{table:s_input_size} and Table.\ref{table:v_input_size}.
We first fix the voxel resolution to 200$\times$200$\times$6, and discretely take different image resolutions from 256$\times$448 to 928$\times$1600 for verification.
The results in Table.\ref{table:s_input_size} show that the increasing in resolution greatly helps to improve the performance of the model, and Fast-BEV achieves the best 0.491 NDS with 832$\times$1440 input image size.

We then fix the image resolution to 384$\times$1056, and try to use different voxel resolutions as shown in Table.\ref{table:v_input_size}. 
We observe that $200\times 200\times 6$ works well for 3D detection, and increasing the resolution from the spatial plane or the height dimension does not help improving the performance.

\textbf{2D/3D Encoder.} To evaluate the performance of different 2D encoders, we perform an ablation study in Table.\ref{table:encoder} (upper). As the encoder grows larger from ResNet-18 to ResNet-101, the mAP and NDS increases by a large margin with over 5\% and 4\%. 
In terms of 3D encoders, as shown in Table.\ref{table:encoder} (lower), when the encoder grows larger from 2 to 6 blocks, the mAP and NDS for detection task increases by 0.1\% and 0.5\% respectively. The scale of the 2D encoder has a greater impact on performance than the 3D encoder.


\textbf{Multi-Scale.} We know that multi-scale features can lead to meaningful performance improvement in 2D image detection task. We introduce multi-scale features into the 3D BEV detection task, and analyze the performance improve-
~\\
~\\
\vspace{129.5mm}

\noindent ment brought by multi-scale features from multi-scale images (Image MS), multi-scale BEV(BEV MS), and multi-scale images\&BEV. As shown in Table.\ref{table:multi-scale}, under the same setting, when Image MS and BEV MS are used separately, performance decreases slightly. When using Image MS and BEV MS at the same time, it can bring 0.3\% mAP and 0.4\% NDS performance improvements. When we increase the training epochs, it can be seen from Table.\ref{tab:val_performance} Fast-BEV(R50)MS setting that MS can bring 0.9\% mAP and 1.2\% NDS, more performance improvements.

\textbf{Epoch.} To investigate the influence of training epochs, we do an ablation study in Table.\ref{table:epoch}.
We observe that baseline and Fast-BEV achieve the best 0.338\% NDS and 0.451\% NDS at epoch 24 and 48, respectively. We find that the upper-bound of Fast-BEV is much higher than the baseline. And Fast-BEV requires more training epochs to achieve better results, because of the strong data augmentation and temporal feature fusion.

\textbf{Augmentation.} As shown in Table.\ref{table:augmentation}, we observe that the performance is significantly improved whether using image augmentation or BEV augmentation alone. 
For image augmentation, mAP and NDS increased by 3.8\% and 2.8\%, respectively. 
For BEV augmentation, mAP and NDS increased by 1.8\% and 2.3\%, respectively. When the two augmentation are used together, the mAP and NDS can be further improved by 4.6\% and 4.4\%. 

\textbf{Multi-Frame Feature Fusion.}
To investigate the effectiveness of multi-frame feature fusion, we show an ablation study in Table.\ref{table:sequantial}. 
When adding one history frame, the mAP and NDS are significantly improved with 2.8\% and 7.8\%. When further increase to four history frames, the mAP and NDS continue to improve by  with 3.0\% and 9.3\%, showing that temporal information is important for 3D detection.

\textbf{Speed Contribution of Fast-Ray components.} Table.\ref{tab:inference_speed_abalation} shows the impact of Multi-View to One-Voxel and LUT on inference speed on CPU and GPU with CUDA platforms respectively. Using the M$^2$BEV setting as the baseline, we gradually added Multi-View to One-Voxel and LUT operations to measure the speed. Multi-View to One-Voxel and LUT of Fast-Ray both contribute more to speed.

\textbf{View Transformation Type.}
Table.\ref{tab:view_transformation_for_performance} shows the performance comparison of just replacing the view transformation under the same training setting. The Fast-Ray type has only a slight decrease in mAP and NDS compared to the baseline (M$^2$BEV) type. This is negligible compared to the significant increase in inference speed it brings.

\begin{table}
    \centering
    \caption{The ablation of LUT and Multi-View to One-Voxel on inference speed.}
    \begin{tabular}{c|c|c}
        \noalign{\smallskip}\hline\noalign{\smallskip}
        Method& CPU (ms) & CUDA (ms)\\
        \noalign{\smallskip}\hline\noalign{\smallskip}
        M$^2$BEV (baseline) &54.0 & 1.2 (2080Ti)   \\
        + Multi-View to One-Voxel & 21.3& 0.16 (2080Ti)  \\
        + LUT& 7.0 & 0.006 (2080Ti)   \\
        \noalign{\smallskip}\hline\noalign{\smallskip}
    \end{tabular}
    \vspace{-1em}
    \label{tab:inference_speed_abalation}
\end{table}
\begin{table}
    \centering
    \caption{The ablation of view transformation performance.}
    \begin{tabular}{c|c|c|c}
        \noalign{\smallskip}\hline\noalign{\smallskip}
        Method& View Transformation Type &mAP & NDS\\
        \noalign{\smallskip}\hline\noalign{\smallskip}
        (1) & M$^2$BEV type&0.337 & 0.478   \\
        (2) & Fast-Ray type&0.334 & 0.473 \\
        \noalign{\smallskip}\hline\noalign{\smallskip}
    \end{tabular}
    \vspace{-1em}
    \label{tab:view_transformation_for_performance}
\end{table}
\begin{table}[!t]
    \centering
    \caption{Latency comparison of view transformation of different methods on CPU and GPU platforms. The image resolution is 256 $\times$ 704. And the device is 2080Ti.}
    \begin{tabular}{c|c|c}
        \noalign{\smallskip}\hline\noalign{\smallskip}
        Model& CPU Latency (ms) & CUDA Latency (ms)\\
        \noalign{\smallskip}\hline\noalign{\smallskip}
        M$^2$BEV &54.0 & 1.2  \\
        LSS &66.8  & 6.6\\
        BEVPool & 66.8 & 1.5 \\
        BEVPoolV2 & 66.8 & 0.2 \\
        Fast-BEV& 7.0 & 0.006   \\
        \noalign{\smallskip}\hline\noalign{\smallskip}
    \end{tabular}
    \vspace{-1.5em}
    \label{tab:view_trans_speed_of_three_methods}
\end{table}
\textbf{Speed Comparison of View Transformation.}
Fast-BEV uses M$^2$BEV as the baseline, and it also compares the speed with the original LSS, improved BEVPool, and the optimal speed solution BEVPoolV2\cite{huang2022bevpoolv2} of the LSS series. We compared the speeds of these methods, as shown in Tab.\ref{tab:view_trans_speed_of_three_methods}. Taken together, Fast-BEV has the optimal inference speed and is significantly faster than other solutions on the CPU.

\subsection{Visualization}
As shown in Fig.\ref{fig:fig_vis}, the row1\&3 are visualizations of the 6 views of the baseline scheme, and the row2\&4 are of \ours. 
\ours uses image\&BEV augmentation, four temporal fusion and multi-scale features, the baseline scheme does not use any improved configuration. From the visualization of detection results, it can be seen that \ours has a significant improvement on distant occluded small objects. This visualization phenomenon is consistent with performance improvement brought by data augmentation, temporal fusion and multi-scale features.

%% file: Sections/benchmark.tex
\begin{table*}[t]
\centering
\caption{The modular design of the serial models denoted as M0-5 are presented. The type of image encoder is chosen from R18/R34/R50. Voxel resolution is denoted as x-y-z for the view transformation from image features to BEV features. The number of block x and channel y denoted as xb-yc are clarified for BEV encoders. M0-5 all use a single scale feature.
}
\setlength{\tabcolsep}{3mm}{
\label{table:benchmark_performance}
\resizebox{0.85\textwidth}{!}{
\begin{tabular}{c|c|c|c|c|c|c|c}
 
\noalign{\smallskip}\hline\noalign{\smallskip}

Model&Name&Image Encoder& Image Res.& Voxel Res.& BEV Encoder & mAP &NDS\\
\noalign{\smallskip}\hline\noalign{\smallskip}
ResNet&M0 & R18 & 256$\times$704 & 200$\times$200$\times$4 & 2b-192c & 0.277 & 0.411  \\
&M1 & R18 & 320$\times$880 & 200$\times$200$\times$4 & 2b-192c &0.295 & 0.421  \\
&M2 & R34 & 256$\times$704 & 200$\times$200$\times$4 & 4b-224c &0.326 & 0.455  \\
&M3 & R34 & 256$\times$704 & 200$\times$200$\times$6 & 6b-256c &0.331 & 0.467  \\
&M4 & R50 & 320$\times$880 & 250$\times$250$\times$6 & 6b-256c & 0.346 & 0.482  \\
&M5 & R50 & 512$\times$1408 & 250$\times$250$\times$6 & 6b-256c & 0.369 & 0.493  \\

\noalign{\smallskip}\hline\noalign{\smallskip}

\end{tabular}
}
}
\end{table*}

\begin{table*}[t]
\centering
\caption{The latency on different platforms (Xavier, Orin, T4) are evaluated with the sum of three parts including an image encoder(2D), view transformation(2D-to-3D) and a BEV encoder(3D). Latency indicates the overall time-consumption and 2D/2D-to-3D/3D indicates the latency breakdown column of the three parts from left to right respectively.
}
\setlength{\tabcolsep}{2mm}{
\label{table:benchmark_latency}
\resizebox{1\textwidth}{!}{
\begin{tabular}{c|cc|cc|cccc}
\toprule 
\multicolumn{1}{c}{} & \multicolumn{2}{c}{Xavier, 22TOPS without DLA} & \multicolumn{2}{c}{Orin, 170TOPS without DLA} & \multicolumn{2}{c}{T4, 130 TOPS}  \\
\cmidrule(r){2-3} \cmidrule(r){4-5} \cmidrule(r){6-7}  
\multicolumn{1}{c}{Name} & 
\multicolumn{1}{c}{Latency(ms)} &\multicolumn{1}{c}{2D/2D-to-3D/3D(ms)} &
\multicolumn{1}{c}{Latency(ms)} & \multicolumn{1}{c}{2D/2D-to-3D/3D(ms)} & 
\multicolumn{1}{c}{Latency(ms)} & \multicolumn{1}{c}{2D/2D-to-3D/3D(ms)} \\
\noalign{\smallskip}\hline\noalign{\smallskip}
M0 & 50.7 (19.7 fps) & 15.2 / $<0.5$ / 35.0    & 22.3 (44.8 fps) & 9.9 / $<0.2$ / 12.2      & 10.3 (97.1 fps) & 4.5 / $<0.1$ / 5.7 \\
M1 & 49.0 (20.4 fps) & 13.4 / $<0.5$ / 35.1    & 24.2 (41.3 fps) & 10.4 / $<0.2$ / 13.6      & 12.2 (82.0 fps) & 6.4 / $<0.1$ / 5.7    \\
M2 & 66.1 (15.1 fps) & 22.7 / $<0.5$ / 42.9    & 23.1 (43.3 fps) & 9.5 / $<0.2$ / 13.4      & 12.2 (82.0 fps) & 5.4 / $<0.1$ / 6.7    \\
M3 & 96.1 (10.4 fps) & 23.0 / $<0.5$ / 72.6    & 31.0 (32.3 fps) & 9.8 / $<0.2$ / 21.0      & 17.9 (55.9 fps) & 5.6 / $<0.1$ / 12.2   \\
M4 & 159.7 (6.3 fps) & 48.8 / $<0.5$ / 110.4    & 47.9 (20.9 fps) & 15.5 / $<0.2$ / 32.2      & 29.0 (34.5 fps) & 10.3 / $<0.1$ / 18.6  \\
M5 & 230.4 (4.3 fps) & 119.5 / $<0.5$ / 110.4    & 60.3 (16.6 fps) & 27.0 / $<0.2$ / 33.1      & 42.5 (23.5 fps) & 23.8 / $<0.1$ / 18.6  \\

\noalign{\smallskip}\hline\noalign{\smallskip}
\end{tabular}
}
} 
\end{table*}
\begin{table}
    \centering
    \caption{Latency results of M0 setting on TI-TDA4VM chip.}
    \resizebox{0.48\textwidth}{!}{
    \begin{tabular}{c|c|c|c}
        \noalign{\smallskip}\hline\noalign{\smallskip}
        2D model(DSP) & 2D-to-3D(CPU) & 3D model(DSP) & Overall \\
        \noalign{\smallskip}\hline\noalign{\smallskip}
         29.5ms&10.5ms&31.7ms& 71.7ms(13.9fps)  \\
        \noalign{\smallskip}\hline\noalign{\smallskip}
    \end{tabular}}
    \label{tab:TI_results}
\end{table}

\section{Benchmark}

\noindent{\textbf{Efficient Model Series.}} 
In order to suffice the deployment requirements of different computing power platforms, we have designed a series of efficient models from M0 to M5, as shown in the Table.\ref{table:benchmark_performance}. We set different image encoders (from ResNet18 to ResNet50), image resolutions (from $256\times704$ to $900\times1600$), voxel resolutions (from $200\times200\times4$ to $250\times250\times6$) and BEV encoders (from 2b-192c to 6b-256c) to design model dimensions. It can be seen from the Table.\ref{table:benchmark_performance} that from M0 to M5, as the image encoder, image resolution, voxel resolution, and BEV encoder gradually become larger, the model performance is gradually increasing.

\noindent{\textbf{Deployment on Popular Devices.}}
In addition to focusing on performance, we deploy the M series models on different on-vehicle platforms (Xavier, Orin, T4) with the acceleration of CUDA-TensorRT-INT8. Specifically, AGX Xavier has a computing power of 22TOPS without DLA acceleration deployed with CUDA11.4-TRT8.4.0-INT8, AGX Orin 64G has a computing power of 170TOPS without DLA acceleration deployed with CUDA11.4-TRT8.4.0-INT8, and T4 has a computing power of 130TOPS deployed with CUDA11.1-TRT7.2.1-INT8. As shown in Table.\ref{table:benchmark_latency}, we evaluate the latency of M series models on these on-vehicle devices, and breakdown the latency into 2D/2D-to-3D/3D three parts. From Table.\ref{table:benchmark_latency}: (1)We can see that as the M series models gradually become larger, the performance is gradually improved, the latency on the same computing platform is also basically increased gradually. And the latency of 2D and 3D parts also increases respectively. (2)From left to right, as the actual computing power\footnote{The nominal computing power of T4 platform is not as good as Orin platform, but the actual computing power is stronger than Orin platform in the M series models.} of these three devices gradually increases, the latency of each model of the M series decreases gradually, and the latency of 2D and 3D parts decreases respectively. (3)Combining the performance of the M series in Table.\ref{table:benchmark_latency}, we can find that when only latency is considered, the M0 model can reach 19.7 FPS on a low-computing platform such as Xavier, which can achieves real-time inference speed. Considering the performance, the M2 model has the most reasonable trade-off between performance and latency. On the premise of being comparable to the performance of the R50 series model in Table.\ref{tab:val_performance}, it can reach 43.3 FPS on the Orin platform, which can achieve the actual real-time inference requirement.

In addition, as shown in Table.\ref{tab:TI_results}, we add the latency measurement results on the Texas Instruments TI-TDA4VM chip. As a vehicle AI chip, its computing power is extremely low, only 8 TOPs, so we only report the latency measurement results of M0 setting of Table.\ref{table:benchmark_performance}. With only 8 TOPs of computing power, it can achieve an inference speed of 13.9 fps, which reflects Fast-BEV’s powerful inference performance on low computing power non-CUDA platforms.

\begin{figure}[t]
    \begin{center}
    \includegraphics[width=1\linewidth]{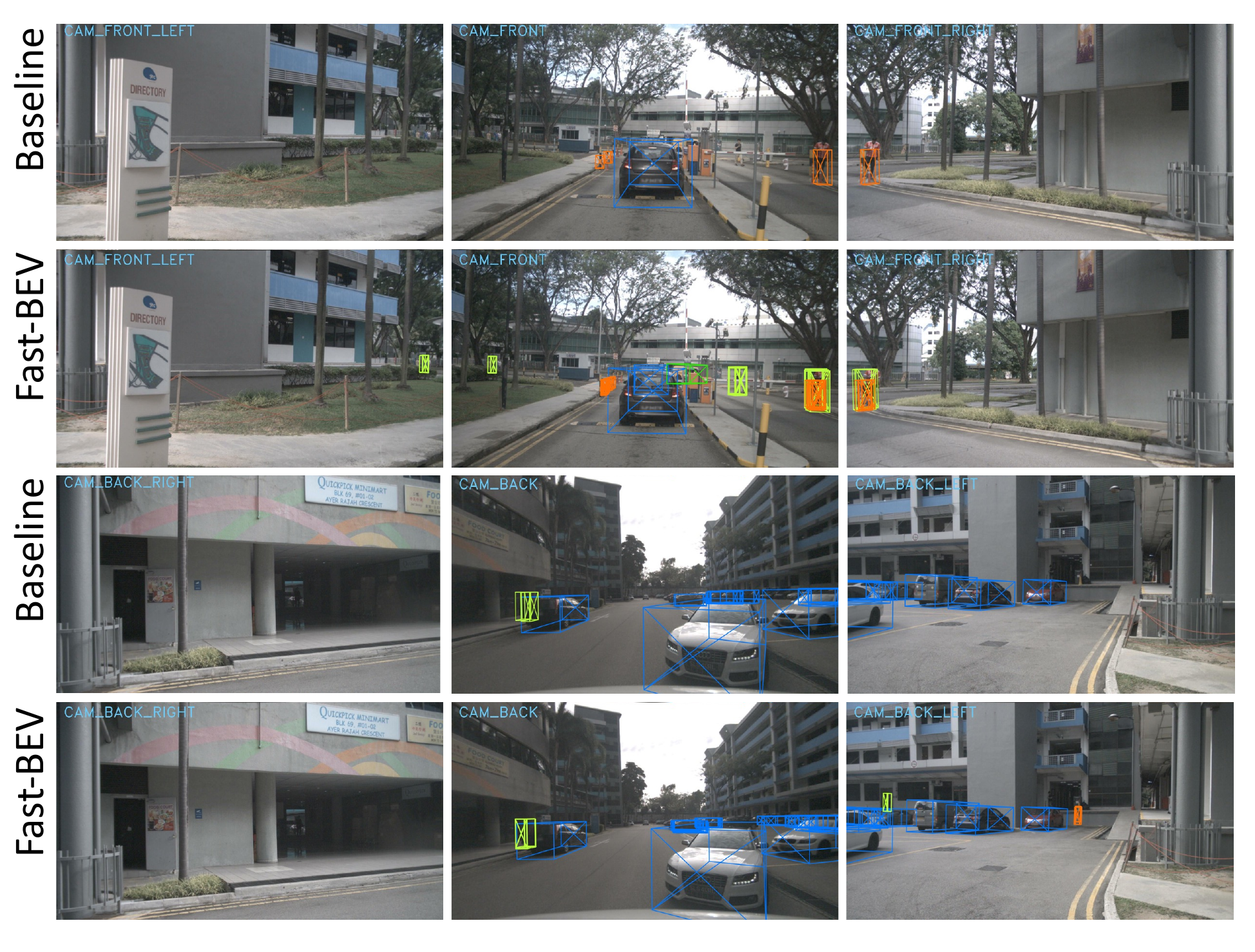}
    \end{center}
    \caption{Visualization comparison of \ours and baseline scheme without any improved
configuration.
    }
\label{fig:fig_vis}
\end{figure}

%% file: Sections/application_discussion.tex
\section{Application Discussion}
At present, there are more and more BEV perception solutions, but they are mainly in the pursuit of academic filed performance, and few of them take into account how to better deploy them on on-vehicle chips, especially low-computing chips. We cannot deny that work such as BEVDet\cite{huang2021bevdet} and BEVDepth\cite{li2022bevdepth} are among the first possible solutions currently considered due to their convenience when deploying on on-vehicle chips. But \ours gives the possibility of application in the following cases.

\begin{itemize}
    \item \textbf{Low-computing power chips.} Although the computing power of autonomous driving chips is gradually increasing, some chips with low computing power, such as Nvidia Xavier, are still used in economical vehicles. \ours can perform better on low computing power chips with faster speed.
    \item \textbf{Non-GPU Deployment.} The successful deployment and application of DEVDepth\cite{li2022bevdepth} and BEVDet\cite{huang2021bevdet} mainly rely on the efficient voxel pooling operation supported by CUDA multi-threading. However, non-GPU chips without CUDA libraries, such as Texas Instruments chips with DSP as the computing unit, are difficult to develop DSP multi-threading operators, and the CPU speed is not fast enough, which make their solutions lose their advantages on such chips. \ours provides convenience of deployment on non-GPU chips with its fast CPU speed.
    
    \item \textbf{Scalable in practice.} With the development of technology, many autonomous manufacturers have begun to abandon lidar and only use pure cameras for perception. 
    There is no depth information in the large amount of data collected by pure cameras in real vehicles. However, in actual development, model scaling up or data scaling up is often based on data collected from real vehicles with these pure camera, in order to leverage data potential for performance improvement. In this case, depth supervision based solutions encounter bottlenecks, while \ours does not introduce any depth information and can be better applied.
\end{itemize}

%% file: Sections/conclusion.tex
\section{Conclusions}
In this paper, we propose \ours, a faster and stronger fully convolutional BEV perception framework which is suitable for on-vehicle deployment. To improve the efficiency and performance, the \ours consists of: a Fast-Ray transformations to fast transfer image feature to BEV space, a multi-scale image encoder to leverage multi-scale information for better performance, an efficient BEV encoder to speed up on-vehicle inference, a strong data augmentation strategy to avoid over-fitting, a multi-frame feature fusion mechanism to leverage the temporal information. 

We hope our work can shed light on the industrial-level, real-time, on-vehicle BEV perception.
\noindent In the future, we are interested in extending \ours with more modalities input, such as LiDAR and RADAR.
We are also interested in supporting more tasks in the \ours system, such as 3D tracking, motion prediction, and HD reconstruction.